\documentclass[11pt]{article}

\usepackage[margin=1in]{geometry}
\usepackage{amsmath}
\usepackage{booktabs}
\usepackage{hyperref}
\usepackage{url}
\usepackage{graphicx}
\usepackage{authblk}

\usepackage{microtype}
\usepackage[numbers]{natbib}

\title{\textbf{Predicting Tennis Serve Directions with Machine Learning}}

\author{
Ying Zhu and Ruthuparna Naikar \\
Georgia State University, Atlanta, USA \\ 
\texttt{yzhu@gsu.edu}, \texttt{rnaikar1@student.gsu.edu}
}

\date{}

\begin{document}
\maketitle
\begin{center}
\begin{minipage}{0.9\textwidth}
\noindent\textbf{Abstract.} 
Serves, especially first serves, are very important in professional tennis. Servers choose their serve directions strategically to maximize their winning chances while trying to be unpredictable. On the other hand, returners try to predict serve directions to make good returns. The mind game between servers and returners is an important part of decision-making in professional tennis matches. To help understand the players' serve decisions, we have developed a machine learning method for predicting professional tennis players' first serve directions. Through feature engineering, our method achieves an average prediction accuracy of around 49\% for male players and 44\% for female players. Our analysis provides some evidence that top professional players use a mixed-strategy model in serving decisions and that fatigue might be a factor in choosing serve directions. Our analysis also suggests that contextual information is perhaps more important for returners' anticipatory reactions than previously thought.

\medskip
\noindent\textbf{Keywords:} Tennis serve directions~$\cdot$ Machine learning~$\cdot$ Prediction.
\end{minipage}
\end{center}

\section{Introduction}

At the beginning of each point in a tennis match, the serving player needs to
decide where to direct the serve. There are three serve directions: wide, body
(to the returner), and down-the-T (to the middle of the court) for each serve.
Each player makes about 100 such decisions in a typical professional tennis
match. These are important decisions because serves, especially first serves,
are crucial in professional tennis matches~\cite{mecheri2016,odonoghue2017,rioult2015}.
A fast and well-placed first serve gives the server a big advantage. For
example, based on ATP statistics~\cite{atp}, Novak Djokovic (world No.~1 for
much of the last seven years) has won 76.5\% of his first service points and
53.4\% of his second service points. In other words, without his first serves,
Djokovic was only slightly better than his opponents.

The player who serves (the \emph{server}) chooses the serve directions
strategically to maximize the winning chances. On the other hand, the player
who returns the ball (the \emph{returner}) will try to predict serve directions
by analyzing serve patterns, which is especially important for the fast first
serves. The average first serve speed is about 180--200~km/h for male tennis
professionals and about 150--170~km/h for female professionals. With less than
0.5 seconds to react to such powerful serves, a returner needs a fast physical
reaction and reasonably accurate anticipation. Previous research suggested that
tennis players' anticipatory responses are informed by both the kinematics of
serve motion and contextual information~\cite{vernon2018,vernon2020} but the
nature of such anticipatory responses is still being debated~\cite{aviles2019}.

The mind game between a server and a returner makes it an interesting case
for studying human decision-making in a highly competitive environment. Some
economists have used game theories to analyze professional players' serve
patterns, with mixed results~\cite{walker2001,hsu2007,spiliopoulos2016,gauriot2020,anderson2021}.
As more tennis data are available, Wei et al.~\cite{wei2015} have developed a
machine learning method to predict serve directions by analyzing the Hawkeye
data, with a prediction accuracy of 27.8\%.

In this paper, we present our work on predicting professional tennis players'
first serve directions by machine learning. Our machine learning model uses
human annotated professional tennis match data, without video analysis or
Hawkeye data. Through feature engineering and model tuning, we achieved an
average prediction accuracy of about 49\% for male players and about 44\% for
female players. Our results provide more insights into the behavior of both
tennis servers and returners. Our analysis provides some evidence that top
professional players use a mixed strategy for choosing serve directions and that
fatigue might affect such decisions. In addition, our work shows that it is
possible to achieve reasonable serve direction predictions solely based on
contextual information, suggesting that contextual information is perhaps more
important for returners' anticipatory reactions than previously thought.

Professional tennis players rarely reveal their on-court decision-making
process. Since many decisions are made intuitively, a player may not even be
aware of their own tendencies and patterns. Building machine learning models to
predict serve directions may help us gain insights into their decision-making
process as well as the differences between players.

Players and coaches may also use the machine learning model to evaluate the
predictability of their serves. Our study shows that some players' serves are
more predictable than others. Some players may be more predictable serving
from the ad side than from the deuce side and vice versa. Such information can
guide players to fine-tune their own games or study their opponents' games.

\section{Related Work}

Tennis data have been used in many academic research projects. Here we focus
on the analysis of tennis serve directions. Some economists have used game
theories to study the optimal strategy for choosing the serve directions.
Walker and Wooders~\cite{walker2001} analyzed ten professional tennis matches
and found that the theory of mixed-strategy Nash equilibrium can largely
explain the top players' selection of serve directions. They noted that top
players tended to switch strategies frequently, resulting in serial dependence
and higher predictability. Hsu et al.~\cite{hsu2007} revisited Walker and
Wooders' work using a broader data set and found no significant evidence of
serial dependence. However, Spiliopoulos~\cite{spiliopoulos2016} analyzed the
data from the Match Charting Project~\cite{sackmann2022} and found some top
male players had higher serial dependencies in serve directions than others.
More recently, Gauriot et al.~\cite{gauriot2020} analyzed the Hawkeye data
from over 3000 tennis matches played at the Australian Open and confirmed the
finding by Walker and Wooders~\cite{walker2001}. However, Anderson et
al.~\cite{anderson2021} analyzed the data from the Match Charting
Project~\cite{sackmann2022} and rejected a key implication of a
mixed-strategy Nash equilibrium, that the probability of winning a service
game is the same for all serve directions. They argued that the dynamic
programming strategy is more efficient than the mixed strategy.

Wei et al.~\cite{wei2015} analyzed the Hawkeye data from three years of the
Australian Open men's draw and developed a method to predict serve directions.
They considered 14 serve directions (seven for the deuce side and seven for the
ad side) and used machine learning techniques (e.g., Random Forest) to make
predictions. Their input parameters are score, player style, and opponent style.
A player's style is the distribution of the player's serve count in the 14
directions. Their highest prediction accuracy is 27.8\%.

Our work is similar to Wei et al.~\cite{wei2015} in that we both try to predict
tennis serve directions using machine learning techniques. The difference is
that we use the Match Charting Project data~\cite{sackmann2022} instead of the
Hawkeye data. We use a different set of features, and we adopt the commonly
used six serve directions (wide, body, and down-the-T for the deuce or ad
side) rather than 14 directions.

Kovalchik and Reid~\cite{kovalchik2018} analyzed the Hawkeye data from singles
matches at the 2015 to 2017 Australian Open and built a taxonomy of shots via
clustering. They reported the overall distributions of serve directions for the
deuce and ad side separately but did not predict serve directions for individual
players.

Tea and Swartz~\cite{tea2022} analyzed the ball tracking data from the 2019
and 2020 Roland Garros tournaments, which contain 82 men's and 81 women's
matches. They used Bayesian Multinomial Logistic Regression to build a
predictive model of serve directions. They found discernible differences between
male and female players and between individual players. Their model can output
predictive distributions of serve directions. An example with Roger Federer's
data was discussed, but the model prediction accuracy for individual players
was not reported.

Whiteside and Reid~\cite{whiteside2016} used machine learning (k-means
clustering) to analyze the Hawkeye data of tennis serves to study the optimal
landing locations for aces. They found three key elements related to serve
aces: direction relative to the returner, closeness to the lines, and speed.
De~Leeuw et al.~\cite{deleeuw2020} used subgroup discovery to find the
characteristics of won service points for a specific professional tennis player.
They found that more points were won if the player avoided hitting a backhand
after the serve. These two studies are not relevant to our work on predicting
serve directions.

Now we will look at serves from a returner's perspective. Returning first
serves in professional tennis is one of the most difficult tasks in sports, and
yet professional players have been able to return most first serves. Many
analysts do not believe that fast physical reaction alone is enough to explain
the many successful returns because the reaction time is less than 0.5~second.
These players must have learned to read the serves with reasonable accuracy.
However, the exact nature of this anticipatory behavior is still
unknown~\cite{aviles2019}. The most widely examined source of anticipatory
information has been the kinematics of serve motion. In other words, a
returner might be able to predict the serve directions from reading the serve
motion before the ball is hit. But professional players are trained to disguise
their serve directions by maintaining the same serve motion. Therefore, reading
the kinematics of serve motion is difficult. Some studies showed that contextual
information might be useful for predicting the serve
directions~\cite{vernon2018,vernon2020}. Our work is related to this subject
because our results suggest that it is possible to make reasonably accurate
predictions solely based on contextual information.

\section{Basic Information about Tennis Serves}

A tennis match is divided into sets, games, and points. A tennis court is
laterally divided into two sides: the \emph{deuce side} and the \emph{ad side}.
The two players take turns serving for each game. The serving player serves
from the deuce side and ad side alternatively.

At the start of each point, the serving player has two chances to serve. If
the first serve fails, the player can make a second serve. A player usually
makes a faster but more risky first serve and a slower but safer second serve.
The player can serve toward anywhere in the service box, but there are generally
three directions: wide, body (toward the returner), and down-the-T (toward the
middle of the court). In this study, we only consider the first serves because
the first serves give the server a significant advantage. Therefore, in the
discussions below, the word ``serve'' means ``first serve'' by default.

\section{Data}

We used the data from the Match Charting Project~\cite{sackmann2022}. This
open-source project provides detailed point-by-point and shot-by-shot data for
thousands of professional tennis matches. Unlike the Hawkeye data, this data
set is created by a group of volunteers watching tennis match videos and
manually entering coded shot-by-shot data, including the serve directions and
outcomes. An Excel script then derives additional information from the
shot-by-shot data, such as the score, who is serving, and rally length for each
point. The human-coded serve direction for each point is the ground truth for
our model training and testing.

The data set we analyzed contains 3424 matches of 655 male players and 1916
matches of 422 female players. However, most players only have one or a few
matches in the database. Therefore, we only run the analysis for a selected
group of players with at least 30 matches.

We processed and analyzed the data using Python-based tools such as
Pandas~\cite{mckinney2010}, Sklearn~\cite{pedregosa2011},
SciPy~\cite{virtanen2020}, etc. The original data set contains errors, such as
missing values in match data, duplicate or incorrect match IDs, match IDs in
the point data set but are not in the match data set, and data entry errors in
some shot-by-shot codes. So we spent a lot of time on data cleaning and
transformation. The original data set primarily contains the scores and
shot-by-shot descriptions for each match.

Several previous works~\cite{spiliopoulos2016,anderson2021} also used the
Match Charting data, but they did not use machine learning.

\section{Feature Engineering}

The performance of a machine learning program depends on the selected features.
We went through an iterative process of extracting, selecting, and testing
features. In addition to the original features in the Match Charting data set,
we also derived many new features from the original point-by-point and
shot-by-shot data set. For example, we calculated the number of serves a
player made toward each direction, identified critical points, calculated how
many shots a player had played before each point, and estimated how much a
player had run in the last point, etc.

Many features were tested and rejected. The features discussed in this paper
are the ones that currently generate the highest prediction accuracy. We are
still working on feature engineering. New features may be added, and some of
the existing features may be modified or removed in the future.

We try to select features likely to influence a player's serve decisions. In
addition to predicting serve directions, we also want to see if certain features
are more important in making such decisions, which may help us gain insights
into the players' decision-making process. We will discuss each of the selected
features in the subsections below.

\subsection{Outcome of Previous Points}

We believe that the outcome of previous points would influence the selection of
serve directions. We assume that a professional player would have a rough idea
of how many points he or she has served toward each direction and how many
points are won. This is the assumption made by several previous
works~\cite{anderson2021,gauriot2020,walker2001}. The analysis by
Spiliopoulos~\cite{spiliopoulos2016} also showed the serial dependency of serve
directions on the previous point's serve direction and outcome.

The following features are calculated and used in our machine learning model.

\begin{itemize}
    \item For each server and each point, our program calculates the count of
    the serves made toward each direction, from the beginning of the match to
    the previous point (3~features). These parameters are similar to the
    ``prior style'' parameters in Wei et al.~\cite{wei2015}, but we only use
    three directions for the deuce or ad side while Wei et al.\ used 7
    directions for each side. Although a player may not be able to remember the
    exact count of serves made toward each direction, the player should have a
    rough idea of the counts for the last several service games. Therefore, a
    variation of this feature is the counts of serve directions from a certain
    number of service games prior to the current point rather than from the
    beginning of the match. But it is not easy to determine how many prior
    service games should be considered.

    \item For each server and each point, our program also calculates the count
    of serves the server made toward each direction and won, from the beginning
    of the match to the previous point (3~features). Similarly, a variation of
    this feature is to count only for a certain number of service games before
    the current point, not from the beginning of the match.

    \item For each point, the program calculates the percentage of ``good''
    first serves the server made toward each direction, from the beginning of
    the match to the previous point (3~features). These are the so-called
    ``serve percentage'' for each direction. A professional player should have a
    reasonably accurate understanding of their current serve percentage.

    \item For each point, the program records the winner of the previous point
    (1~feature).
\end{itemize}

\subsection{Fatigue}

As the match progresses, both players become more and more tired. We want to
examine whether the level of fatigue is a factor in choosing serve directions.
It is reasonable to assume that a player will exploit the opponent's fatigue in
choosing serve directions. For example, serving wide is likely to make an
opponent run more because a wide serve opens up the court more. The player's
own fatigue may also affect serve directions. Because the net is lower in the
middle, serving to the T may require less jumping.

The following features are used to estimate fatigue in our machine learning
model.

\begin{itemize}
    \item For each point, our program calculates the cumulative run indexes for
    two players from the beginning of the match to the previous point
    (2~features). They indicate how tired each player is from the running and
    hitting before each serve.
\end{itemize}

Because the Match Charting data~\cite{sackmann2022} contains detailed
shot-by-shot information, including the shot type (e.g., forehand, backhand,
slice, volley, overhead), shot direction (i.e., to-deuce-side, to-middle, or
to-ad-side), and the depth of each shot (e.g., shallow and deep), our program
can infer the player's court position when they hit a particular shot. Based on
that information, the program can estimate how much a player ran for each point
and calculate a ``run index.'' It is more accurate than the shot (rally) count
because it includes running. This run index is not as accurate as the Hawkeye
data for measuring running distance, but it is a consistent estimate from point
to point. Since Hawkeye data are not publicly available, it is difficult to get
more accurate measures.

\subsection{Performance Anxiety}

Because a tennis match does not have a time limit, scoreboard pressure is the
primary source of a player's performance anxiety. Such anxiety could influence
a player's decisions. For this reason, Wei et al.~\cite{wei2015} used scores in
their machine learning model. But our method is different. Instead of using
scores, we calculate an index of each player's performance anxiety. Our work is
based on the OCC model~\cite{occ1988} of emotion, which is the standard model
in affective computing. Based on the OCC model, anxiety is influenced by hope,
fear, and uncertainty.

A player's feeling of uncertainty is related to the gap between the two
players' scores. The smaller the gap, the higher the uncertainty. If the score
is tied, the uncertainty is the highest. If one player is very close to winning,
the uncertainty is very low. Due to tennis' hierarchical score structure, there
are three levels of uncertainty. The gap between the set scores influences the
match-level uncertainty. The gap between the game scores influences the
set-level uncertainty. The gap between the point scores influences the
game-level uncertainty.

A player's feeling of hope depends on how close the player is to winning. If
a player's score is close to the winning score, the player's hope is high.
Again, there are three levels of hope: match-level hope, set-level hope, and
game-level hope.

A player's feeling of fear depends on how close the player is to losing. There
are three levels of fear: match-level fear, set-level fear, and game-level fear.

For example, if a player leads by a significant margin and serves for the set
point, the player's hope is high, fear is low, and uncertainty is low, resulting
in a relatively low level of anxiety. On the other hand, if the scores are very
close near the end of a match, such as in the final set tiebreak, each player
will have high hope, high fear, and high uncertainty, resulting in high levels
of anxiety for both players.

In our model, a performance anxiety index is calculated separately on the
game, set, and match level based on the following equation (3~features):
\begin{equation}
    \mathit{performance\ anxiety} = \mathit{uncertainty}  * (\mathit{hope} + \mathit{fear})
\end{equation}

The overall performance anxiety is the sum of game, set, and match level
anxiety indices (1~feature):
\begin{equation}
    \mathit{overall\ anxiety} = \mathit{game\ anxiety} + \mathit{set\ anxiety} + \mathit{match\ anxiety}
\end{equation}

In the equations, we do not consider fear to be the negative of hope.
Therefore, fear does not reduce hope, and vice versa. This is because hope and
fear can coexist in most situations. For example, in a close tiebreak game, a
player is close to both victory and defeat at the same time. In such cases,
both strong hope and strong fear can coexist.

\subsection{Other Features}

We also considered other factors that may influence the serve decisions, such
as court surfaces~\cite{hughes2021} and the opponent's handedness. For example,
a player might serve differently against a lefty opponent.

\section{Machine Learning}

We ranked the players by the number of matches they have in the data set and
analyzed players with at least 30 matches in the data set. Due to the space
limit, we only present the results for ten male players and ten female players.
The players are selected based on their current ranking and significant
achievements. But the results for other players are generally consistent with
those presented here.

We applied the following machine learning models to our data set: Multinomial
Logistic Regression, Decision Tree, Random Forest, Support Vector Machine
(Multiclass Classification), and Neural Network. We also applied Bagging
classifier, Ada Boost classifier, and XGBoost classifier, but the results are
no better than the models mentioned above, so we do not present their results
here.

For Random Forest, we used 200 trees with a maximum depth of 150. For the
Bagging classifier, we used 50 estimators. For the Ada Boost classifier, we
used 70 estimators. For the XGBoost classifier, the $K$ value is 10. For the
neural network, we use \texttt{sklearn.neural\_network.MLPClassifier()} with
two hidden layers (200, 100).

We train our models individually for each selected player. For each player, we
randomly split the data into a training set (70\%) and a testing set (30\%), and
we use the same training set and testing sets for the different machine learning
models. For each player, our program selects all the points that this player
serves, and predicts the first serve direction using the features discussed in
Section~5. The prediction is then compared with the actual first serve direction
coded by the person who entered the data for the Match Charting project. The
prediction accuracy is calculated based on all the points in the testing data
set.

The results of the analysis are presented in
Tables~\ref{tab:men_deuce}--\ref{tab:women_ad}. In the tables, LR stands for
Multinomial Logistic Regression, DT stands for Decision Tree, RF stands for
Random Forest, SVM stands for Support Vector Machine, and NN stands for Neural
Network.

\begin{table}[ht]
\centering
\caption{First Serve Direction Prediction Accuracy for the Deuce Side Serves (men)}
\label{tab:men_deuce}
\begin{tabular}{llcccccr}
\toprule
First name & Last name & LR & RF & DT & SVM & NN & MEAN \\
\midrule
Novak      & Djokovic  & 0.47 & 0.50 & 0.45 & 0.46 & 0.47 & 0.47 \\
Roger      & Federer   & 0.46 & 0.49 & 0.44 & 0.47 & 0.47 & 0.47 \\
Nick       & Kyrgios   & 0.55 & 0.52 & 0.50 & 0.55 & 0.54 & 0.53 \\
Daniil     & Medvedev  & 0.45 & 0.49 & 0.48 & 0.47 & 0.46 & 0.47 \\
Andy       & Murray    & 0.56 & 0.53 & 0.58 & 0.52 & 0.54 & 0.55 \\
Rafael     & Nadal     & 0.52 & 0.50 & 0.43 & 0.52 & 0.52 & 0.50 \\
Dominic    & Thiem     & 0.55 & 0.47 & 0.43 & 0.55 & 0.55 & 0.51 \\
Stefanos   & Tsitsipas & 0.49 & 0.48 & 0.46 & 0.47 & 0.47 & 0.47 \\
Stan       & Wawrinka  & 0.46 & 0.49 & 0.45 & 0.45 & 0.48 & 0.46 \\
Alexander  & Zverev    & 0.46 & 0.42 & 0.42 & 0.47 & 0.46 & 0.44 \\
\midrule
\multicolumn{2}{l}{MEAN} & 0.50 & 0.49 & 0.46 & 0.49 & 0.50 & 0.49 \\
\bottomrule
\end{tabular}
\end{table}

\begin{table}[ht]
\centering
\caption{First Serve Direction Prediction Accuracy for the Ad Side Serves (men)}
\label{tab:men_ad}
\begin{tabular}{llcccccr}
\toprule
First name & Last name & LR & RF & DT & SVM & NN & MEAN \\
\midrule
Novak      & Djokovic  & 0.49 & 0.50 & 0.44 & 0.49 & 0.48 & 0.48 \\
Roger      & Federer   & 0.56 & 0.52 & 0.53 & 0.59 & 0.56 & 0.55 \\
Nick       & Kyrgios   & 0.50 & 0.49 & 0.45 & 0.48 & 0.48 & 0.48 \\
Daniil     & Medvedev  & 0.53 & 0.51 & 0.53 & 0.49 & 0.60 & 0.53 \\
Andy       & Murray    & 0.48 & 0.45 & 0.44 & 0.48 & 0.47 & 0.46 \\
Rafael     & Nadal     & 0.54 & 0.51 & 0.45 & 0.54 & 0.54 & 0.52 \\
Dominic    & Thiem     & 0.58 & 0.54 & 0.52 & 0.58 & 0.58 & 0.56 \\
Stefanos   & Tsitsipas & 0.47 & 0.45 & 0.42 & 0.48 & 0.47 & 0.46 \\
Stan       & Wawrinka  & 0.49 & 0.49 & 0.43 & 0.49 & 0.47 & 0.47 \\
Alexander  & Zverev    & 0.45 & 0.47 & 0.44 & 0.44 & 0.44 & 0.45 \\
\midrule
\multicolumn{2}{l}{MEAN} & 0.51 & 0.49 & 0.47 & 0.51 & 0.51 & 0.50 \\
\bottomrule
\end{tabular}
\end{table}

\begin{table}[ht]
\centering
\caption{First Serve Direction Prediction Accuracy for the Deuce Side Serves (women)}
\label{tab:women_deuce}
\begin{tabular}{llcccccr}
\toprule
First name  & Last name  & LR & RF & DT & SVM & NN & MEAN \\
\midrule
Victoria    & Azarenka   & 0.47 & 0.47 & 0.40 & 0.49 & 0.48 & 0.46 \\
Ashleigh    & Barty      & 0.51 & 0.45 & 0.45 & 0.51 & 0.48 & 0.48 \\
Angelique   & Kerber     & 0.37 & 0.37 & 0.33 & 0.40 & 0.36 & 0.36 \\
Anett       & Kontaveit  & 0.41 & 0.43 & 0.39 & 0.39 & 0.39 & 0.40 \\
Garbine     & Muguruza   & 0.47 & 0.45 & 0.40 & 0.47 & 0.45 & 0.45 \\
Naomi       & Osaka      & 0.49 & 0.41 & 0.41 & 0.48 & 0.45 & 0.45 \\
Karolina    & Pliskova   & 0.44 & 0.45 & 0.40 & 0.43 & 0.44 & 0.43 \\
Maria       & Sakkari    & 0.44 & 0.44 & 0.37 & 0.47 & 0.47 & 0.44 \\
Iga         & Swiatek    & 0.41 & 0.43 & 0.36 & 0.43 & 0.44 & 0.41 \\
Serena      & Williams   & 0.48 & 0.49 & 0.44 & 0.49 & 0.49 & 0.48 \\
\midrule
\multicolumn{2}{l}{MEAN} & 0.45 & 0.44 & 0.40 & 0.46 & 0.45 & 0.44 \\
\bottomrule
\end{tabular}
\end{table}

\begin{table}[ht]
\centering
\caption{First Serve Direction Prediction Accuracy for the Ad Side Serves (women)}
\label{tab:women_ad}
\begin{tabular}{llcccccr}
\toprule
First name  & Last name  & LR & RF & DT & SVM & NN & MEAN \\
\midrule
Victoria    & Azarenka   & 0.46 & 0.41 & 0.34 & 0.47 & 0.43 & 0.42 \\
Ashleigh    & Barty      & 0.51 & 0.46 & 0.45 & 0.46 & 0.46 & 0.47 \\
Angelique   & Kerber     & 0.61 & 0.58 & 0.48 & 0.61 & 0.60 & 0.58 \\
Anett       & Kontaveit  & 0.43 & 0.42 & 0.42 & 0.43 & 0.42 & 0.42 \\
Garbine     & Muguruza   & 0.39 & 0.40 & 0.38 & 0.37 & 0.40 & 0.39 \\
Naomi       & Osaka      & 0.44 & 0.38 & 0.36 & 0.45 & 0.47 & 0.42 \\
Karolina    & Pliskova   & 0.43 & 0.42 & 0.42 & 0.46 & 0.44 & 0.43 \\
Maria       & Sakkari    & 0.50 & 0.51 & 0.45 & 0.50 & 0.53 & 0.50 \\
Iga         & Swiatek    & 0.49 & 0.40 & 0.37 & 0.49 & 0.43 & 0.44 \\
Serena      & Williams   & 0.47 & 0.47 & 0.45 & 0.47 & 0.48 & 0.47 \\
\midrule
\multicolumn{2}{l}{MEAN} & 0.47 & 0.44 & 0.41 & 0.47 & 0.47 & 0.45 \\
\bottomrule
\end{tabular}
\end{table}

We also calculated the serve direction distributions for each player. Our
results are similar to those reported by Tea and Swartz~\cite{tea2022}. The
serve direction distributions vary from player to player. For example,
Djokovic's serve directions are more evenly distributed, while Federer tended
to serve much less to the body.

\section{Discussion}

From Tables~\ref{tab:men_deuce} and~\ref{tab:men_ad}, we can see that our
machine learning models achieved an average 49\% prediction accuracy for the
deuce side serve directions and 50\% accuracy for the ad side serve directions
for the selected male players. Adding other male players will bring the average
percentage slightly lower to around 48\%. From Tables~\ref{tab:women_deuce}
and~\ref{tab:women_ad}, we can see that our machine learning models achieved an
average 44\% prediction accuracy for the deuce side serve directions and 45\%
accuracy for the ad side serve directions for the selected female players.
Adding other female players will bring the average percentage slightly lower to
around 43\%. From the tables, we can also see that prediction accuracy is
generally consistent among different machine learning methods.

We found only one published work by Wei et al.~\cite{wei2015} that reported a
serve direction prediction accuracy (27.8\%). However, it is difficult to
compare our prediction accuracy with theirs because Wei et al.\ used seven
serve directions per side while we used the more traditional three directions
per side. This is because we based our analyses on different ground truths.
Wei et al.\ used the Hawkeye data as ground truth, and they could divide the
serve directions into smaller groups. We used human-observed serve directions
as our ground truth, and our data only has three serve directions per side. The
features we used are also quite different from the features used by Wei et al.

We conducted a feature importance analysis for the Decision Tree model. We
found that the most important features are the cumulative counts of first serves
made to each direction, the run index of the server in the previous point, and
the first serve percentage for each direction. While the prediction accuracy
varies for each player, these three features are consistently among the most
important. This provides some indirect evidence that they might also be the
important factors a player considers when choosing serve directions. The
importance of cumulative counts of first serve directions and first serve
percentage are consistent with the mixed-strategy findings by Walker and
Wooders~\cite{walker2001}, Spiliopoulos~\cite{spiliopoulos2016}, and Gauriot et
al.~\cite{gauriot2020}. But as far as we know, the importance of server fatigue
(run index) in choosing the serve direction has not been discussed in previous
work.

Finally, our results show that it is possible to achieve reasonably accurate
prediction of serve directions solely based on contextual information such as
the outcome of previous serves, performance anxiety, and fatigue. This may help
explain why professional players are able to return most of the very fast first
serves. Many analysts do not believe that fast physical reaction alone is enough
to explain the many successful first serve returns because a returner must react
to a first serve in less than 0.5~seconds. These players must have learned to
read the serves with reasonable accuracy. Although the exact nature of this
anticipatory behavior is still unclear~\cite{aviles2019}, the most widely
examined source of anticipatory information has been the kinematics of serve
motion. In other words, a returner might be able to predict the serve directions
from reading the serve motion before the ball is hit. But professional players
are trained to disguise their serve directions by maintaining the same serve
motion, making it difficult to read. Some studies showed that contextual
information might be useful for predicting the serve
directions~\cite{vernon2018,vernon2020}. Our work suggests that contextual
information is perhaps more important for returners' anticipatory reactions than
previously thought.

\section{Conclusion and Future Work}

We have described our machine learning methods for predicting professional
tennis players' first serve directions. Through feature engineering, our method
achieves an average prediction accuracy of around 49\% for male players and
44\% for female players.

Our feature importance analysis provides some indirect evidence that the top
professional players seem to use a mixed-strategy model in choosing serve
directions, which is consistent with some previous
works~\cite{walker2001,spiliopoulos2016,gauriot2020}. However, the importance
of server fatigue in choosing the serve direction has been a new discovery. Our
work also suggests that contextual information is perhaps more important for
returners' anticipatory reactions than previously thought.

We are continuing our work on feature engineering to improve prediction
accuracy. We will also test using Brier Score as a measurement of prediction
accuracy. We also plan to apply our method to applications in other highly
competitive situations.

\bibliographystyle{unsrtnat}

\end{document}